\documentclass[11pt]{article}

\usepackage[preprint]{acl}

\usepackage{times}
\usepackage{latexsym}

\usepackage[T1]{fontenc}

\usepackage[utf8]{inputenc}

\usepackage{microtype}

\usepackage{inconsolata}

\usepackage{graphicx}
\usepackage{booktabs}

\usepackage{booktabs,tabularx,multirow,array}
\newcolumntype{Y}{>{\raggedright\arraybackslash}X}

\title{Evaluating Medical LLMs by Levels of Autonomy: A Survey Moving from Benchmarks to Applications}

\author{
  Xiao Ye$^{1}$ ~~~~ Jacob Dineen$^{1}$ ~~~~ Zhaonan Li$^{1}$ ~~~~ Zhikun Xu$^{1}$ ~~~~ Weiyu Chen$^{1}$\\
  \textbf{Shijie Lu$^{1}$ ~~~~ Yuxi Huang$^{1}$ ~~~~ Ming Shen$^{1}$ ~~~~ Phu Tran$^{2}$ ~~~~ Ji\textendash{}Eun Irene Yum$^{2}$}\\
  \textbf{Muhammad Ali Khan$^{2}$ ~~~~ Muhammad Umar Afzal$^{2}$ ~~~~ Irbaz Bin Riaz$^{2}$ ~~~~ Ben Zhou$^{1}$}\\[0.25em]
  $^{1}$Arizona State University \quad $^{2}$Mayo Clinic\\
  \texttt{xiaoye2@asu.edu}
}

\begin{document}
\maketitle
\begin{abstract}
Medical Large language models achieve strong scores on standard benchmarks; however, the transfer of those results to safe and reliable performance in clinical workflows remains a challenge. This survey reframes evaluation through a levels-of-autonomy lens (L0–L3), spanning informational tools, information transformation and aggregation, decision support, and supervised agents. We align existing benchmarks and metrics with the actions permitted at each level and their associated risks, making the evaluation targets explicit. This motivates a level-conditioned blueprint for selecting metrics, assembling evidence, and reporting claims, alongside directions that link evaluation to oversight. By centering autonomy, the survey moves the field beyond score-based claims toward credible, risk-aware evidence for real clinical use.
\end{abstract}

\section{Introduction}

\begin{figure*}[t]
  \centering
  \includegraphics[width=\linewidth]{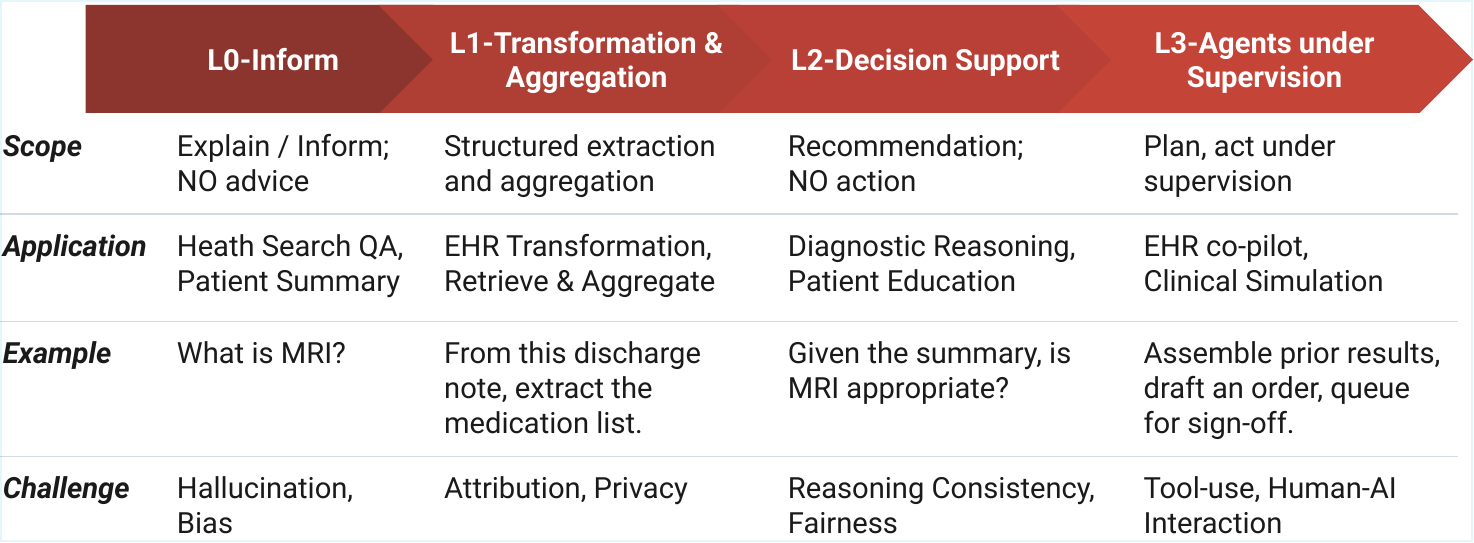}
  \caption{Overview of autonomy levels (L0--L3) for medical LLMs, showing for each level the scope, typical applications, an example question/task, and key challenges. The rightward arrow indicates increasing autonomy and risk.}

  \label{fig:survey-teaser}
  \vspace{-1\baselineskip}
\end{figure*}

Large language models (LLMs) have advanced rapidly on medical benchmarks \citep{singhal2023large,hendrycks2020measuring,Nazi2024LLMHealthcareReview}. Both general-purpose \citep{yang2025qwen3, deepseekai2025deepseekr1incentivizingreasoningcapability, openai_introducing_gpt5_2025} and domain-specialized \citep{singhal2025medpalm2, bolton2024biomedlm27bparameterlanguage} LLMs now achieve high scores on licensing-style examinations and medical Q\&A benchmarks, and they perform well on clinical summarization tasks \citep{Oliveira2025ClinicalNoteSumm, Tang2023MedEvidenceSummarization}. These headline results suggest that LLMs could meaningfully assist clinicians and patients across a range of information-centric workflows.

However, benchmark correctness alone is not sufficient for clinical use \citep{Hager2024LLMLimitsClinDecision, ma2025leaderboardrethinkingmedicalbenchmarks}. Clinical deployment requires consistency, fairness, auditability, calibrated uncertainty, and demonstrably safe clinical reasoning \citep{Omiye2023RaceBased,Fehr2024Transparency}. Most benchmarks are Q\&A-centric, which rarely probe these aspects, allowing unsafe reasoning and missing context to go undetected \citep{Soroush2024PoorCoders,asgari2025framework}. A rigorous multidimensional evaluation that spans factual grounding, reasoning quality, uncertainty calibration, safety, and human preferences is required \citep{Tam2024QUEST,shool2025systematic}.

This survey reviews the state of LLM evaluation in the medical field and identifies its limitations. We first summarize how LLMs are being applied in the medical field and what current benchmarks measure. While these scores are informative, they only provide quick capability snapshots and overlook integration into real workflows, calibration, and traceable provenance. Therefore, we move from scores to applications, treating evaluation as a means of showing that a system is sufficient for a defined purpose and scope at a specific autonomy level. For each level, we define the scope, typical applications, evaluation focus, boundaries, and challenges: \textbf{L0} Inform (no personalized advice); \textbf{L1} Information Transformation \& Aggregation (structure, summarize); \textbf{L2} Decision Support (recommendations and personalized advice); and \textbf{L3} Agents Under Human Supervision (plan + invoke tools/APIs to enqueue actions under explicit review). This organization makes evaluation targets explicit: factual grounding at L0/L1; calibrated reasoning and coverage at L2; tool-use safety and auditability at L3. Regarding the challenges in each level, they are cumulative, not isolated: each higher autonomy level inherits unresolved issues from lower levels. As autonomy and permitted actions expand, new risks emerge that are specific to that level’s capabilities. Finally, we outline future work and recommendations for developing more robust evaluation frameworks to ensure that LLM-based systems can be trusted in clinical practice.

\section{Related Work}

Contemporary medical-LLM surveys mostly list datasets (e.g., USMLE/OKAP), task scenarios, and evaluation modes, but they rarely organize evaluation targets by autonomy level or permitted clinical actions. Representative examples include a review contrasting closed and open-ended tasks and discussing agentic settings \citep{chen2025imed}, and a systematic review highlighting reliance on general-purpose models and accuracy metrics with limited calibration and safety assessment \citep{shool2025systematic}. Parallel strands propose conversation-quality and safety evaluations \citep{abbasian2024foundation}, human-rater rubrics such as QUEST \citep{Tam2024QUEST}, multi-dimensional criteria in SCORE \citep{tan2024score}, and reporting guidance in TRIPOD-LLM \citep{gallifant2025tripodllm}. We address this gap by mapping evaluation objectives and metrics to autonomy levels (L0–L3). This mapping clarifies what evidence is sufficient for a system’s intended role, which risks must be tested and where human oversight is required. It also provides a practical blueprint for selecting datasets, metrics, and the identification of possible risks for different clinical actions.

Autonomy scales outside medical-LLM evaluation exist but target system design or human factors, not evaluation blueprints. In clinical decision research, levels of autonomy delineate who acts and who bears responsibility \citep{festor2021autonomy}. In the broader agent literature, five-level design frameworks define autonomy via user roles (operator $\rightarrow$ observer) \citep{feng2025levels}, and industry taxonomies describe L0–L5 agentic behavior \citep{vellum2025levels}. None of these prescribe autonomy-conditioned metrics or align evaluation with healthcare oversight and risk. Our survey fills that gap by integrating autonomy scales with concrete measurement choices for medical LLMs, so assessment can evolve with the agent’s permitted actions and governance requirements.

\section{Evaluation Methodologies}
\label{sec:evaluation}
We start by listing current benchmarks and metrics for medical LLMs. This sets clear measurement boundaries before we move to the applications. For LLMs in the medical field, benchmarks provide low-cost, repeatable evidence about specific capabilities within a bounded scope; they surface failure modes early, track progress over time, and help align claims with a system’s intended role (L0–L3). First, we group benchmarks by task and summarize the usefulness of each and the L0–L3 level(s) they inform (\S\ref{subsec:task_benchmarks}). We then summarize automated and human metrics and their limitations (\S\ref{subsec:metrics}). The list here is not exhaustive; a complete, expanded table appears in Appendix~\ref{subsec:full_table}.

\subsection{Benchmarks}
\label{subsec:task_benchmarks}

\noindent\textbf{Exam Q\&A.}
\textbf{Task definition:} answer multiple-choice or short-answer questions derived from professional medical examinations across specialties.
\textbf{Typical benchmarks:} MedQA (USMLE) \citep{jin2021disease}; MedMCQA (Indian medical entrance exams) \citep{pal2022medmcqa}.
\textbf{Usefulness:} low-cost unit tests of factual breadth and specialty coverage that gate L0–L1 informational tools and surface coarse knowledge gaps.

\noindent\textbf{Summarization.}
\textbf{Task definition:} generate faithful summaries, patient-friendly simplifications, or structural rewrites of clinical text.
\textbf{Typical benchmarks:} MS$^{2}$ (multi-document evidence summaries) \citep{wang-etal-2022-overview}; PubMed long-document summarization \citep{cohan-etal-2018-discourse}.
\textbf{Usefulness:} evaluates whether the output is faithful to the source (no omissions or fabrications) and structurally complete for explanation tasks (L0–L1).

\noindent\textbf{Retrieval-augmented generation (RAG).}
\textbf{Task definition:} answer questions grounded in retrieved documents with explicit attribution to evidence passages.
\textbf{Typical benchmarks:} BEIR (heterogeneous zero-shot retrieval) \citep{thakur2021beir}; TREC-COVID (pandemic literature search) \citep{roberts2021treccovid}.
\textbf{Usefulness:} demonstrates that responses are supported by cited sources and do not contradict them, enabling freshness checks on evolving topics (L0–L1).

\noindent\textbf{Information extraction.}
\textbf{Task definition:} extract and normalize clinical entities, relations, and codes from notes and reports.
\textbf{Typical benchmarks:} CBLUE (NER and diagnosis normalization) \citep{zhang2021cblue}
\textbf{Usefulness:} establishes reliable structuring and normalization to support downstream aggregation in L1 systems.

\noindent\textbf{Decision Support.}
\textbf{Task definition:} make thresholded recommendations or triage decisions in vignettes or multi-turn clinical scenarios scored by clinician rubrics
\textbf{Typical benchmarks:} HealthBench (multi-turn, physician-authored rubrics) \citep{arora2025healthbench}; MedHELM (holistic medical evaluation suite) \citep{bedi2025medhelm}.
\textbf{Usefulness:} targets selective reliability for L2 systems by quantifying calibrated behavior at deployable thresholds and surfacing harm-proximal errors in simulation.

\noindent\textbf{Clinical Dialogue.}
\textbf{Task definition:} Conduct multi-turn conversations with patients or clinicians to achieve task goals while communicating safely and clearly.
\textbf{Typical benchmarks:} MedDialog (EN/ZH) \citep{zeng2020meddialog}; MedDG (ZH) \citep{liu2022meddg}.
\textbf{Usefulness:} assesses communication effectiveness and safety across L0–L2.

\subsection{Automated vs.\ Human Metrics}
\label{subsec:metrics}

\noindent\textbf{Automated Metrics.}
Automated metrics quantify answer correctness, calibration, faithfulness, and retrieval quality. For Q\&A, Exact Match and F1 summarize accuracy, while ECE, Brier, and NLL capture probability calibration (ECE compares confidence to accuracy; Brier is the mean-squared error of predicted probabilities; NLL penalizes over-confident errors) \citep{guo2017calibration,brier1950,manning2008ir}. In selective prediction, risk-coverage curves relate error to the answered fraction and justify abstention on uncertain cases \citep{geifman2017selective,geifman2019selectivenet,traub2024overcoming}. For summarization, lexical overlap (ROUGE/chrF) and embedding-based similarity (BERTScore/BLEURT) are commonly complemented with source-grounded checks for omissions and contradictions \citep{lin2004rouge,popovic2015chrf,zhang2020bertscore,sellam2020bleurt}. In RAG settings, retrieval ranking is assessed with Recall@k, MRR, and nDCG \citep{manning2008ir}. In practice, care is needed: ECE can be sensitive to binning and mis-rank models unless robust estimators or verified calibration are used \citep{nixon2019measuring,kumar2019verified}; risk-coverage summaries are hard to compare across tasks without principled area measures \citep{traub2024overcoming}; overlap/embedding metrics correlate only weakly with factuality, motivating explicit factuality/entailment auditing \citep{maynez-etal-2020-faithfulness}; and strong retrieval alone does not ensure correct answers, with automatic attribution evaluation remaining challenging \citep{li2024attributionbenchhardautomaticattribution,joren2025sufficientcontextnewlens}.

\noindent\textbf{Human Evaluation.}
Clinician raters typically apply behaviorally anchored rubrics (BARS-style) to score clinical correctness, coverage, contextualization, reasoning transparency, uncertainty handling, readability, actionability/safety, and empathy; inter-rater agreement is summarized with Cohen’s $\kappa$ or ICC \citep{holland2022bars,mchugh2012kappa}. For patient-facing text, readability targets are checked with Flesch/Flesch--Kincaid or SMOG indices \citep{singh2024readability,badarudeen2010readability}. Reporting should also acknowledge known issues: $\kappa$ can appear low despite high raw agreement when class prevalence or bias shifts \citep{feinstein1990high,cicchetti1990resolving}; and readability formulas may disagree by several grade levels on the same passage and primarily capture surface difficulty rather than genuine comprehension, so multiple indices and task-specific checks are advisable \citep{wang2013readability}.

\noindent\textbf{LLM-as-Judge (LAJ).}
To scale open-ended assessments, many studies adopt LLM-as-judge: strong models render pairwise preferences or rubric-based scores, often with prompts that request quoted evidence. To control bias, studies randomize order or blind positions and, when feasible, ensemble multiple judges. \citep{zheng2023llmasjudge,li2024llmsasjudges,gu2024survey-laj,zhu2025judgelm,tan2025judgebench}. LAJ can track human preferences well in aggregate but is vulnerable to position and verbosity biases and self-preference effects; accordingly, order randomization, style/length normalization, and regular human spot-checks are recommended \citep{chen2024humans-judge-bias,ye2025justice-llm-judge,wataoka2024selfpref}.

\section{Applications and Autonomy Levels}
\emph{From Scores to Applications.}
Static benchmarks provide helpful, quick overviews of the model capabilities, but they do not capture real workflow context, calibration and abstention, provenance and audit trails, or role-specific needs. We therefore treat evaluation as showing that a system is \textbf{sufficient for a defined purpose and scope} at a given autonomy level. Concretely, each autonomy level is organized into five parts: (1) \textbf{Definition \& scope}; (2) \textbf{Typical applications}-representative real tasks; (3) \textbf{Evaluation focus}-what to test and report; (4) \textbf{Scope boundaries}-what the level does not cover; and (5) \textbf{Challenges}. Benchmarks remain useful as ingredients in this assessment, not the destination.

\subsection{L0 - Inform}

\noindent\textbf{Definition and Scope.}
At autonomy level L0, the system functions purely as an informational tool: it explains medical concepts and provides a general background in plain language. It neither tailors advice to an individual patient nor initiates clinical decisions. Outputs are educational in nature and include an explicit non-advice disclaimer.

\noindent\textbf{Typical Applications.}
Representative L0 tasks include answering common health questions (e.g., “What is MRI?”), producing lay summaries of technical passages, and simplifying lab reports. Public datasets used as proxies include consumer Q\&A corpora and patient-facing summarization sets: HealthSearchQA (commonly searched lay questions) within the MultiMedQA \citep{singhal2023large} suite, the TREC 2017 LiveQA Medical task \citep{abacha2017overview}, PubMedQA for abstract-level research comprehension, and consumer-question summarization datasets such as MeQSum and MEDIQA’21 \citep{abacha2019meqsum,mediqa2021}.

\noindent\textbf{Evaluation Focus.}
L0 evaluation should primarily focus on accuracy, completeness (covering key points), and readability (appropriate and well-organized) \citep{srinivasan2025recaptransparentinferencetimeemotion}. Some benchmarks also include structured rubrics or provide reference contexts that enable additional checks: HealthBench \citep{arora2025healthbench} grades free-text answers with physician-written rubrics, MultiMedQA reports human ratings for long-form consumer answers, and PubMedQA uses the abstract as the reference context. We reference these only to motivate the axes here, not to require source citation as a core L0 practice.

\noindent\textbf{Scope Boundaries.}
L0 outputs are intentionally non-patient-specific and involve minimal reasoning: they recall and lightly synthesize facts but do not conduct case workups, triage, or make recommendations (those belong to L2 Decision Support). As a result, L0 content may omit person-specific contraindications or time-sensitive context and should avoid language that could be interpreted as advice. Readability should target plain-language norms (e.g., consumer-health instruments) to reduce misunderstanding, and common safety disclaimers may be shown without tailoring. Some L0 benchmarks include rubrics or reference contexts for scoring (e.g., physician-written criteria or abstract-based contexts), but we do not treat source citation as a core requirement for L0.

\noindent\textbf{Challenges.}
\textbf{Hallucination:} Even at L0, models often produce fluent but unfaithful text: Broad surveys separate hallucinations into intrinsic and extrinsic, link them to training/decoding choices and weak grounding, and recommend source- and task-aware evaluations instead of generic overlap scores \citep{ji2023hallucination}. In summarization specifically, human studies show that systems invent unsupported details and that standard n-gram metrics miss these errors, motivating faithfulness-oriented checks instead \citep{maynez-etal-2020-faithfulness}. For patient-facing summaries, clinical audits advise using medical rubrics that check whether each claim is supported by the underlying notes and whether important facts are missing-rather than relying on readability alone. \citep{asgari2025framework}. However, because these tools audit rather than eliminate hallucinations, the problem persists. \noindent\textbf{Bias:} We use bias to mean systematic tendencies that push outputs away from truth or intended scope (e.g., agreement pressure, overconfidence, or sensitivity to superficial cues), such that the same input intent can yield subtly distorted explanations. These tendencies plausibly arise from the pretraining objective (next-token prediction on skewed web corpora) and are further shaped by alignment procedures that optimize against human preferences \citep{ouyang2022training,sharma2023sycophancy,xu2025towthoughtswordsimprove, shen2025bowreinforcementlearningbottlenecked}. In L0 applications, such biases surface as assistants mirroring user beliefs rather than correcting them \citep{sharma2023sycophancy} and shifting outputs under small, meaning-preserving prompt changes such as formatting tweaks, option order in multiple-choice settings, or early anchoring hints \citep{sclar2023promptformatting,pezeshkpour2023order,lou2024anchoring, zhou2024conceptualunbiasedreasoninglanguage, li2024deceptivesemanticshortcutsreasoning,ye2025cclearncohortbasedconsistencylearning,li2024famicomdemystifyingpromptslanguage,rrv2025thinktuninginstillingcognitivereflections}. They persist in practice because preference-optimized objectives and prompt conventions are integral to how LLMs are used \citep{kadavath2022know,steyvers2025calibration}. Consequently, mitigations such as reporting performance ranges across prompt formats, simple option-order calibrations, anchor-aware templates, and a conservative tone can reduce bias at L0, but they do not reliably remove it. \citep{sclar2023promptformatting,pezeshkpour2023order,lou2024anchoring,steyvers2025calibration}.

\subsection{L1 - Information Transformation \& Aggregation}
\noindent\textbf{Definition and Scope}
This stage turns raw, heterogeneous clinical data into standardized, computable representations and then combines them with external evidence to produce grounded outputs. In practice, health systems map local EHR fields to an interoperability standard (e.g., HL7 FHIR) or a research schema (e.g., OMOP CDM), attach machine-readable provenance for auditability, and operate over de-identified corpora such as MIMIC-IV that illustrate the target tables (encounters, labs, medications) and privacy constraints \citep{abacha2021overview,alsentzer2023zero,zhang2024multiple,liu2021medical}.

\noindent\textbf{Typical Applications.}
\textbf{EHR data transformation:}
Typical pipelines mix schema harmonization (FHIR/OMOP ETL), clinical NLP to extract entities/attributes from notes, concept normalization to standard vocabularies, and de-identification. i2b2/n2c2 shared tasks supply widely used de-identified note sets for de-identification, concept extraction, relation labeling, and medication-change context, enabling objective measurement of span-level and mapping accuracy \citep{li2016bc5cdr,abacha2021overview,nowak2023transformer,mahajan2023n2c2,henry2019n2c2}. \textbf{Retrieve \& Aggregate:}
On top of the transformed corpus, systems index structured facts (problem lists, meds, labs) and unstructured notes, then pair them with external sources (guidelines, reviews) via retrieval-augmented generation (RAG). Retrieval metrics (Recall@k, nDCG, MRR) assess whether the right evidence is fetched; generation metrics (faithfulness/attribution, grounded-answer rate) check that outputs rely on retrieved passages rather than model priors \citep{cohan2020specter,zhang2024multiple,Tang2023MedEvidenceSummarization}. Representative resources include BEIR for generalizable retrieval evaluation, TREC-COVID for high-stakes, rapidly evolving topics, and domain-specific retrievers such as MedCPT for biomedical search \citep{thakur2021beir,roberts2021treccovid,jin2023medcpt}.

\noindent\textbf{Evaluation Focus.}
\textbf{Transformation}: report extraction/normalization scores (e.g., span-level F1, concept mapping accuracy), coverage/completeness of key fields, and lineage completeness. \textbf{Retrieval \& Aggregation}: report Recall@k/nDCG/MRR; grounded-answer and attribution rates to retrieved passages; contradiction-to-source; and selective prediction/abstention rates under uncertainty \citep{alsentzer2023zero,abacha2021overview,zhang2024multiple}.

\noindent\textbf{Scope Boundaries.}
L1 improves structure, traceability, and access to evidence, but it does not perform the patient-specific reasoning required for diagnosis, test selection, or treatment trade-offs. Real clinical workups must integrate temporality, comorbidities, contraindications, and uncertainty-capabilities not captured by schema or Recall@k alone. Studies also show that (i) retrieval can surface conflicting or outdated sources (e.g., during pandemics) and (ii) LLMs may misattribute or over-trust citations; recent medical RAG evaluations highlight these gaps even when retrieval quality is strong \citep{roberts2021treccovid,peng2023study,hueber2023quality,golan2023chatgpt,dhanvijay2023performance,sezgin2023clinical,team2024gemma,chen2023master}. Thus, L0+L1 should be paired with higher-level (L2+) decision-focused assessments before deployment.

\noindent\textbf{Challenges.}
\textbf{Attribution:} Attribution concerns whether aggregated outputs actually rely on and are traceable to the retrieved sources. Automated audits in the medical domain report that models often cite papers that are only loosely relevant or that do not fully support the generated claims \citep{wu2025sourcecheckup}. Even with explicit attribution metrics and fine-grained factuality scoring, models can pass retrieval tests while still weaving in priors or mixing multiple sources in ways that obscure provenance \citep{yang2025raghealth}. Techniques like Self-RAG improve on-demand retrieval and self-critique, yet do not eliminate misattribution when sources disagree, are low quality, or when prompts nudge the model toward fluent synthesis over faithful quotation \citep{asai2023selfrag,jung2024familiarityawareevidencecompressionretrievalaugmented}. Thus, attribution still remains as a challenge. \textbf{Completeness:} Completeness addresses whether transformed corpora and their aggregations cover all clinical facts (problems, meds, labs, temporality, context) without omissions. Comparative studies find that concept-recognition tools are inconsistent and often miss negations, misread abbreviations, and struggle with ambiguity or misspellings, which leads to missing or distorted facts downstream \citep{lossioventura2023concept}. Beyond extraction, mapping text spans to standard concepts is fragile because benchmark datasets contain many ambiguous terms and do not align well with UMLS coverage, so the ‘correct’ code is often unclear from the beginning \citep{newmangriffis2021ambiguity}. In practice, long EHR narratives, events spread across notes, and uneven local coding leave gaps that retrieval can’t fill when the structured substrate is incomplete. These issues persist because gold standards underrepresent edge cases, annotation guidelines vary, and many L1 evaluations emphasize span-level F1 or Recall@k over end-to-end coverage of clinically critical fields \citep{lossioventura2023concept}. \textbf{Privacy \& lineage:} It evaluates whether L1 transformations are governable and safe to share. De-identification of clinical notes reduces direct identifiers but does not preclude membership inference against downstream models; moreover, LLM-generated synthetic notes can carry similar privacy risks when they closely match real data utility \citep{sarkar2024deidnotenough}. At the model layer, training data extraction and related attacks demonstrate that LLMs can memorize and regurgitate snippets of their training corpora \citep{carlini2021extracting}. These realities make rigorous lineage essential: machine-readable provenance (what data, which ETL/normalizers, which retrievers/rankers) should be recorded using established schemas so that organizations can audit and reproduce outputs \citep{provo2013,mitchell2019modelcards,gebru2021datasheets}. Yet provenance remains challenging in practice because pipelines are multi-hop and frequently updated; components are swapped or fine-tuned; and evidence bases evolve. Without disciplined, standard documentation, downstream users cannot reliably trace how a particular claim was produced. \citep{yang2025raghealth}.

\subsection{L2 - Decision Support}

\noindent\textbf{Definition and Scope.}
At autonomy level L2, the system provides patient-specific recommendations that can assist clinical decision making. By design, L2 depends on upstream EHR transformation and retrieval/aggregation (L1) but adds reasoning over the individual’s data and clinical context.

\noindent\textbf{Typical Applications.}
\textbf{Diagnostic reasoning:} it reads a patient’s history, exam, labs, and imaging to propose differentials with brief rationales and possible next steps; evaluations commonly use vignette-based case sets and prompting schemes that elicit stepwise clinical reasoning \citep{goh2024llm_diagnostic_reasoning_jama_open,savage2024diagnostic}. \textbf{Medication decision support:} it integrates a patient’s active medications, allergies, problems, and recent labs to surface potential adverse drug events, drug–drug interactions, and dosing/contraindication checks; widely used resources include the 2018 n2c2 ADE shared task (concept extraction, relation classification, and end-to-end pipelines) and the SemEval-2013 DDIExtraction task for literature-based DDI detection \citep{henry2019n2c2,segura2013ddiextraction}. \textbf{Patient education.} Patient education tools support patient–clinician communication by turning medical information into clear, usable messages and scaffolding two-way conversations. Typical functions include generating plain-language explanations of conditions, tests, and procedures; prompting patients to ask key questions; and using teach-back so patients restate key points to confirm understanding  \citep{ahrq_pemat,cdc_cci,ahrq_teachback_tool,ihi_askme3,ipdas_site,stacey2021ipdas}. Taken together, these tasks are L2 because they require reasoning over an individual patient’s context to generate recommendations or tailored explanations.

\noindent\textbf{Evaluation Focus.}
For L2 decision support, evaluation should foreground three axes beyond L0’s accuracy/completeness/readability: reasoning consistency, calibration and abstention, and safety. For reasoning, score the process, not just final answers: stepwise diagnostic logic and evidence use should be judged against clinician-authored rubrics or case rationales. For calibration, report reliability at the case level, plus selective prediction curves (risk–coverage) with a tunable “don’t know / escalate” option. For safety, it is essential to track contraindication and guideline-violation rates.

\noindent\textbf{Scope Boundaries.}
At L2, systems give decision support by combining a patient’s EHR context with external evidence to produce advisory recommendations. However, they do not plan tasks, call tools or APIs, or change the record, so a clinician must review and decide. This preserves clinical authority but forces clinicians to turn text into orders and messages, which adds workload and risks transcription errors. These limits motivate a shift to agentic L3 configurations that keep a human in the loop while reducing cognitive burden.

\noindent\textbf{Challenges.}
\textbf{Reasoning consistency \& faithfulness:} It concerns whether patient-specific answers are stable across prompt phrasings and whether the rationales actually support the recommendation. Comparative guideline evaluations show that changing formats or instructions can swing model outputs and agreement, underscoring prompt sensitivity in medical settings \citep{wang2024consistency}. Even when stepwise prompts are used to elicit reasoning, studies find that the generated “explanations” can be plausible yet unfaithful to the features that truly drove the prediction \citep{turpin2023unfaithful,madsen2024faithful,kuang2025atomicthinkingllmsdecoupling,rrv2025thinktuninginstillingcognitivereflections}. In diagnostic contexts, structured prompting can improve transparency but does not guarantee faithful causal grounding of the final answer \citep{savage2024prompt,dineen2025qalignaligningllmsconstitutionally}. The result in practice is that two seemingly careful L2 prompts may yield different plans with rationales that read well but do not reliably reflect the model’s decision process, which keeps consistency and faithfulness an open problem for deployment \citep{turpin2023unfaithful}. \textbf{Confidence calibration:} Here the question is whether stated or implicit confidence tracks correctness so that uncertain cases can be flagged or deferred. Recent medical evaluations show that simple proxies (e.g., log-probabilities) correlate only weakly with error and that token-probability or sampling-based methods improve ranking of uncertainty yet still leave pockets of overconfidence \citep{bentegeac2025token}. Semantic-entropy approaches detect confabulations and better prioritize abstention, but they often require multiple generations or added computation, and their reliability varies by task and domain \citep{farquhar2024semantic,kossen2024sep,pennydimri2025uncertainty}. In real L2 use, these trade-offs mean systems may sound certain on incorrect recommendations or abstain too rarely on edge cases \citep{bentegeac2025token,farquhar2024semantic,feng2025birdtrustworthybayesianinference}. \textbf{Fairness:} The core issue is whether recommendations generalize equitably across patient subgroups and clinical contexts. Specialized benchmarks and audit tools show that biases appear in long-form answers and clinical recommendations that plain accuracy scores don’t catch \citep{pfohl2024equitymedqa}. Purpose-built bias benchmarks for clinical LLMs report subgroup-linked shifts in outputs, while triage studies using counterfactual tests reveal intersectional differences across sex and race \citep{zhang2024climb,lee2025triage}. A recent scoping review highlights uneven coverage across medical fields and limited clinician-in-the-loop evaluation-leaving blind spots in where and how biases manifest \citep{liu2025scoping}. Consequently, subgroup reliability remains a persistent challenge \citep{pfohl2024equitymedqa,liu2025scoping}.

\subsection{L3 - Agents under human supervision}
\noindent\textbf{Definition and Scope.}
We define L3 agents as systems that plan and invoke tools/APIs to initiate actions in clinical workflows while keeping a clinician explicitly “in the loop” for review, modification, and sign-off. Core capabilities are task planning, retrieval, and safe tool use (e.g., querying/reading/writing to EHR endpoints), with human oversight enforced at key checkpoints (e.g., order “draft” states, queued messages/referrals).

\noindent\textbf{Typical Applications.}
\textbf{Clinical Copilot:} clinical copilots plan tasks, fetch chart context, and then draft orders, messages, referrals, or care plans via tool/API calls, pausing for human sign-off-e.g., constellation designs that split a patient-facing agent from specialist agents (Polaris), pharmacy verification agents that stage indication/dose/interaction checks (Rx Strategist), and EHR copilots that navigate records and place draft orders (Almanac Copilot) \citep{mukherjee2024polaris,van2024rx_strategist,zakka2024almanac_copilot}.  \textbf{Sandboxed Simulation:} It embeds agents in controlled clinics to probe plan–act–check loops with audit trails-benchmarks like AgentClinic (multimodal encounters with tool use), AI Hospital (multi-agent roles), ClinicalLab (multi-department diagnostics), and 3MDBench (telemedicine dialogue with assessor agents) \citep{schmidgall2024agentclinic,fan2024aihospital,yan2024clinicallab,sviridov2025_3mdbench,yue2025relatesimleveragingturningpoint}. \textbf{Operation \& EHR Automation:} It coordinates non-diagnostic workflows (e.g., prior authorization) and return proposed actions for approval (RxLens; multi-agent prior-auth pipelines) \citep{jagatap2025rxlens}. In contrast to L2 (textual recommendations), these L3 systems invoke tools to initiate actions but keep a clinician in the loop for review and sign-off.

\noindent\textbf{Evaluation Focus.}
At L3, evaluation shifts from answer quality to supervised action quality: studies should first demonstrate end-to-end task success under human oversight. Second, they must verify tool-use correctness: every API call, order, and parameter matches clinical intent. Finally, they should require auditability via machine-readable provenance of data, prompts, models, tools, parameters, and approvals, leveraging standards such as FHIR \texttt{Provenance} and \texttt{AuditEvent} so that actions can be traced for post-hoc review and governance.

\noindent\textbf{Scope Boundaries.}
L3 agents are limited to draft-and-queue actions under explicit human oversight: they may plan tasks and call approved tools/APIs to prepare orders, messages, referrals, or documentation, but execution requires a clinician’s independent review. Looking ahead, closed-loop system deployments can allow agents to autonomously execute pre-approved, low-risk steps when explicit policies and fine-grained access controls are satisfied, rather than requiring per-action approval.

\noindent\textbf{Challenges.}
\textbf{Tool-use failures:} These arise when an agent plans correctly but issues the wrong API call or constructs malformed parameters, so the action fails even if the reasoning was sound. Recent EHR-agent benchmarks show that agents often get confused by FHIR’s linked resources and multi-step queries, and many actions still fail because the API calls they issue are invalid or not FHIR-conformant \citep{jiang2025fhiragentbench}. This persists because health data are heterogeneous and nested, vendor FHIR implementations vary, and small prompt or formatting shifts can flip tool behavior with little semantic validation in the loop. \textbf{Human–AI interaction dynamics:} These refer to how clinicians review and sign off on queued actions. Studies of decision support show overreliance on automated suggestions and alert fatigue, where users ignore or over-accept system outputs, producing oversight gaps even when accuracy is reasonable \citep{khera2023automationbias,abdelwanis2024automationbias}. The problem endures because busy workflows, long sessions, and uneven interface cues make it hard to calibrate attention and to sustain critical review at every step. \textbf{Auditability and accountability:} These require a traceable record of what data were accessed, which tools were called with what parameters, and who approved the final action. Standards such as FHIR \texttt{AuditEvent} and \texttt{Provenance} define the necessary primitives, and governance frameworks emphasize logging and traceability for post-hoc review \citep{hl7_auditevent,hl7_provenance,nist_rmf_2023,who_lmm_2025}. Yet multi-tool, multi-service agent stacks often lack end-to-end lineage across steps, so reconstructing a failure or near-miss remains difficult in practice.
\section{Future Work}
\noindent\textbf{Closed-looped System}
We expect clinical deployments to shift from single-model helpers to closed-loop, hospital-scale systems composed of cooperating, role-specialized agents that escalate to clinicians at predefined gates \citep{schmidgall2024agentclinic, borkowski2025multiagent}. We call for simple, auditable protocols for handoff, disagreement resolution, and safety gating of tool use, preceded by simulation and shadow deployments. We recommend reporting operational outcomes—deferral rates, near-misses, and governance practices—so the community can compare architectures and converge on safe patterns \citep{consortai2020}.

\noindent\textbf{Guarantees.}
We call for reframing “good” performance around risk-controlled selectivity: in clinical settings, systems should act only when a target risk can be met and otherwise defer, evaluated by risk–coverage rather than average accuracy. In this agenda, reliable high accuracy on a subset with explicit deferral (e.g., 99\% accuracy on 20\% task) is preferable to broad moderate accuracy (e.g., 80\% on 80\%), because the former enables out-of-scope detection and smooth human routing while the latter obscures which cases are wrong. We recommend that future evaluations report calibrated confidence on the acted-on subset, and coverage at target risk with slice breakdowns \citep{guo2017calibration}.

\section{Conclusion}
This survey reframes medical LLMs evaluation around levels of autonomy (L0–L3) so that what a system is allowed to do matches the evidence required to trust it. We emphasize risk coverage rather than average accuracy: safe systems act only when they can meet a target risk and defer otherwise. At L0–L1, the focus is factual fidelity, bias, and structural correctness with clear grounding to sources. At L2–L3, the bar rises to calibrated uncertainty, selective answering, subgroup robustness, tool-use correctness, and verified human oversight.

Taken together, this level-conditioned lens turns benchmark scores into credible, clinically relevant claims. Evaluations are most persuasive when they make the level explicit, pair target risk with achieved coverage, report performance across slices and shifts, and verify both confidence calibration and oversight checkpoints. We hope this provides a clear language for building medical LLMs that are not only capable, but reliably useful—and safe—in practice.
\section*{Limitations}
This survey is necessarily selective and time‐bounded; model releases, datasets, and guidance evolve rapidly, so some details may become outdated. Evidence in the literature remains uneven: many studies emphasize lab benchmarks over prospective or randomized evaluations, and reporting quality is inconsistent.

\bibliography{custom}

\appendix
\section{Appendix}
\label{sec:appendix}

\subsection{Full Table}
\label{subsec:full_table}
\begin{table*}[t]
\footnotesize
\setlength{\tabcolsep}{3pt}\renewcommand{\arraystretch}{0.98}
\centering
\caption{Expanded benchmarks by task class. Each row lists what the task can establish for clinical readiness, common evaluation metrics, the target level (L0--L3), and representative benchmarks.}
\label{tab:expanded_task_benchmarks}
\begin{tabularx}{\textwidth}{@{}l Y Y c Y@{}}
\toprule
\textbf{Task} & \textbf{Usefulness} & \textbf{Metric} & \textbf{Level} & \textbf{Benchmark} \\
\midrule
Knowledge recall / exam Q\&A
& Breadth of factual recall and clinical reasoning on multiple-choice or open-ended exam questions across specialties and difficulty levels. Evaluates recall under constrained formats and coverage gates, and whether models hallucinate when unsure \citep{arora2025healthbench}.
& Exact match/F1; accuracy; subject\slash difficulty slices; contamination checks; calibration on unanswerable or uncertain questions.
& L0
& \textit{MedMCQA} \citep{pal2022medmcqa}; \textit{MultiMedQA} (MedQA, MedMCQA, PubMedQA, MMLU) \citep{singhal2023large}; \textit{Mirage RAG suite (MMLU-Med, MedQA-US, PubMedQA, BioASQ Y\slash N)} \citep{cartwright2024mirage} \\
\addlinespace
Summarization / transformation
& Fidelity and completeness of summaries or transformations of clinical notes, conversations, or research literature. Tests whether models can produce coherent, structurally complete summaries without omissions or hallucinations \citep{deyoung2021ms2,wu2024mimicivbhc}.
& Hallucination/omission rate; ROUGE/BERTScore/chrF; section completeness; clinical correctness; expert ratings.
& L1
& \textit{MS$^2$} \citep{deyoung2021ms2}; \textit{MIMIC-IV-BHC} \citep{wu2024mimicivbhc}; \textit{MTS-Dialog} \citep{bojai2023mtsdialog}; \textit{ACI-Bench} \citep{ravi2023acibench} \\
\addlinespace
Retrieval-augmented QA
& Attribution and faithfulness of answers to retrieved sources and freshness/recency of information. Evaluates how well models retrieve and ground answers in relevant documents \citep{cartwright2024mirage}.
& Faithfulness/attribution; source-contradiction rate; Recall@k, nDCG, MRR; answer correctness; freshness.
& L0,L1
& \textit{Mirage RAG benchmark (MMLU-Med, MedMCQA, PubMedQA, BioASQ)} \citep{cartwright2024mirage}; \textit{HealthSearchQA (part of MultiMedQA)} \citep{singhal2023large} \\
\addlinespace
Evidence-based fact-checking
& Reliability of claims and ability to verify or refute medical statements using evidence. Useful for ensuring LLM outputs do not propagate misinformation.
& Claim classification accuracy; evidence recall/precision; F1 for true/false/unfounded labels; citation quality.
& L0,L1
& \textit{MedFact} \citep{tang2025medfact} \\
\addlinespace
Information extraction / coding
& Structured accuracy on entity recognition, relation extraction, coding, and normalization tasks. Establishes ability to extract structured data from unstructured texts \citep{lu2023biored,li2016bc5cdr,uzuner2011i2b2,zhang2021cblue}.
& Mention/cluster F1; relation F1; coding/normalization accuracy; entity-linking accuracy.
& L1
& \textit{BioRED} \citep{lu2023biored}; \textit{BC5CDR} \citep{li2016bc5cdr}; \textit{n2c2 2010 (i2b2)} \citep{uzuner2011i2b2}; \textit{CBLUE} \citep{zhang2021cblue} \\
\bottomrule
\end{tabularx}
\end{table*}

\begin{table*}[t]\ContinuedFloat
\footnotesize
\setlength{\tabcolsep}{3pt}\renewcommand{\arraystretch}{0.98}
\centering
\caption{Expanded benchmarks by task class (continued).}
\begin{tabularx}{\textwidth}{@{}l Y Y c Y@{}}
\toprule
\textbf{Task} & \textbf{Usefulness} & \textbf{Metric} & \textbf{Level} & \textbf{Benchmark} \\
\midrule
Decision support / triage (simulation)
& Selective reliability for clinical decision making: calibration at deployable thresholds, risk--coverage trade-offs, harm proxies, and quantitative reasoning. Includes simulation of triage, diagnosis, personalized diabetes management, and medical calculations \citep{arora2025healthbench,bedi2025medhelm,arora2024dexbench,zhou2024erreason,lin2024medcalcbench}.
& ECE; Brier; NLL; risk--coverage curves; contraindication/near-miss rates; accuracy, groundedness, safety, clarity, actionability; MAE for calculations.
& L2
& \textit{HealthBench} \citep{arora2025healthbench}; \textit{MedHELM} \citep{bedi2025medhelm}; \textit{DexBench} \citep{arora2024dexbench}; \textit{ER-Reason} \citep{zhou2024erreason}; \textit{MedCalc-Bench} \citep{lin2024medcalcbench} \\
\addlinespace
Clinical dialogue
& Communication quality and human factors in multi-turn doctor--patient conversations or simulated OSCE interviews. Measures goal completion, uncertainty marking, empathy, and adherence to safety rails; also covers note generation from visit dialogues \citep{zeng2020meddialog,ravi2023acibench,koh2024osce}.
& Goal completion; uncertainty/hedging tags; rubric-based ratings; guideline-contradiction flags; empathy/communication scores; note-generation quality; ROUGE/BERTScore.
& L2,L3
& \textit{MedDialog} (Chinese/English) \citep{zeng2020meddialog}; \textit{MTS-Dialog} \citep{bojai2023mtsdialog}; \textit{ACI-Bench} \citep{ravi2023acibench}; \textit{OSCE simulated interview dataset} \citep{koh2024osce} \\
\addlinespace
Multimodal (imaging + text)
& Linkage between radiology images and free-text reports or classification labels; evaluates image understanding, report generation, and cross-modal retrieval \citep{johnson2019mimic,wang2017nihchestxray}.
& Report correctness; finding detection/linking; classification accuracy; precision/recall/F1; bounding-box/segmentation metrics.
& L2,L3
& \textit{MIMIC-CXR}, \textit{MIMIC-CXR-JPG} \citep{johnson2019mimic}; \textit{NIH ChestX-ray} \citep{wang2017nihchestxray} \\
\addlinespace
Patient retrieval
& Ability to retrieve relevant literature or similar patient summaries to support clinicians. Tests retrieval quality and ranking of semantically similar patients or articles \citep{lakshminarayanan2023pmcpatients}.
& Recall@k; nDCG; MRR; patient-similarity accuracy; retrieval precision.
& L3
& \textit{PMC-Patients} \citep{lakshminarayanan2023pmcpatients} \\
\addlinespace
Molecular / drug discovery
& Validity and diversity of generated molecules and optimization of proxy properties for research settings.
& Validity/diversity/novelty proxies; property-optimization success; synthetic accessibility; logP and QED scores.
& L3
& \textit{MOSES} \citep{polykovskiy2020moses}; \textit{GuacaMol} \citep{brown2019guacamol} \\
\bottomrule
\end{tabularx}
\end{table*}

\subsection{LLM Usage}

We used LLMs for language polishing and organization only; all technical claims and citations were authored and verified by the authors

\end{document}